\documentclass{article}
\usepackage{spconf,amsmath,graphicx}

\usepackage{amsfonts}
\usepackage{color}
\usepackage{multirow}

\title{Incorporating supervised domain generalization into data augmentation}

\name{Shohei Enomoto, Monikka Roslianna Busto, Takeharu Eda}
\address{NTT Software Innovation Center}
\begin{document}
%
\maketitle
\begin{abstract}
With the increasing utilization of deep learning in outdoor settings, its robustness needs to be enhanced to preserve accuracy in the face of distribution shifts, such as compression artifacts. 
Data augmentation is a widely used technique to improve robustness, thanks to its ease of use and numerous benefits.
However, it requires more training epochs, making it difficult to train large models with limited computational resources.
To address this problem, we treat data augmentation as supervised domain generalization~(SDG) and benefit from the SDG method, contrastive semantic alignment~(CSA) loss, to improve the robustness and training efficiency of data augmentation.
The proposed method only adds loss during model training and can be used as a plug-in for existing data augmentation methods.
Experiments on the CIFAR-100 and CUB datasets show that the proposed method improves the robustness and training efficiency of typical data augmentations.
\end{abstract}
\begin{keywords}
Data Augmentation, Feature Alignment, Robustness, Supervised Domain Generalization
\end{keywords}
\section{Introduction}
\label{sec:intro}
Deep learning has become increasingly practical in outdoor settings like autonomous driving and smart cities in recent years. 
However, these use cases face distribution shifts, which decrease the accuracy of deep neural networks, due to sensor noise, blurring, compression artifacts, etc.  
To ensure reliable results, the robustness of deep neural networks against such corruptions needs to be enhanced.

There are many studies to improve robustness, of which data augmentation is the most widely used technique.
Despite its simplicity, data augmentation has various benefits in addition to improved robustness, such as improved confidence calibration and transferability.
The high robustness of recent deep neural network architectures such as vision transformer is known to be largely due to data augmentation~\cite{bai2021transformers}.
However, data augmentation increases training complexity, requiring more training epochs than usual and making it difficult to train large models with limited computational resources~\cite{tied_aug}.

To address this problem, we revisit the role of data augmentation, which improves the recognition accuracy for out-of-distribution data through transformations that increase the diversity of the data.
Such a transformation changes the distribution of the data to the one which is different from the source distribution.
In other words, we can view data augmentation as a method to generate labeled out-of-distribution data.

\begin{figure}[t]
\centering
\includegraphics[width=1.0\linewidth]{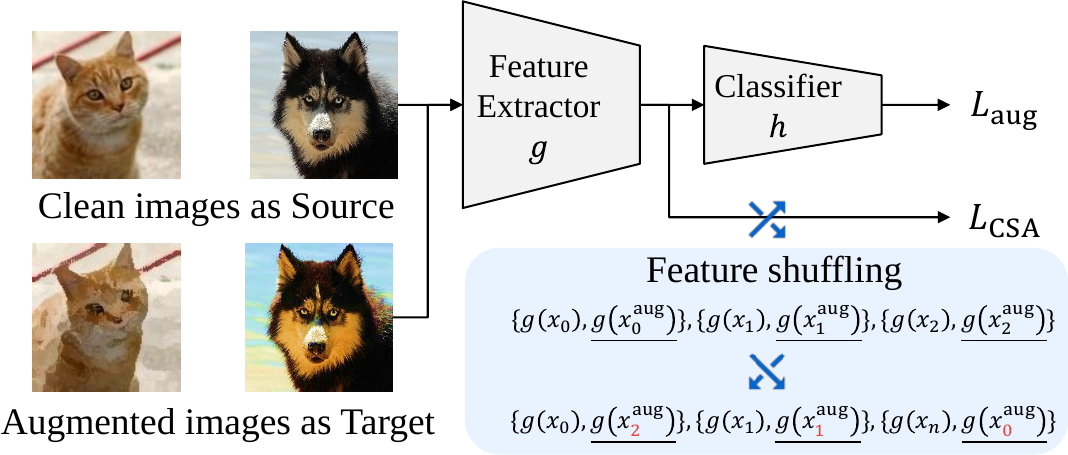}
\caption{Overview of the proposed method. The proposed method introduces CSA loss by treating the clean images as the source distribution images and the augmented images as the out-of-distribution images. Feature shuffling randomly shuffles feature indices to make feature pairs with the same and different labels, making CSA loss applicable.}
\label{fig:overview}
\end{figure}

Based on the observation above, we treat data augmentation as \textit{supervised domain generalization~(SDG)}, with clean data as source distribution data and augmented data as out-of-distribution data.
Data augmentation can benefit from improved accuracy and training efficiency via SDG methods.
In this paper, we introduce the SDG method, contrastive semantic alignment~(CSA) loss~\cite{csa}, into data augmentation.
CSA loss encourages deep neural networks to acquire domain-invariant representations by mapping features with the same labels closer together and separating features with different labels.
It does not lose the simplicity of data augmentation, but it further improves robustness and training efficiency.
In the data augmentation process, pairs of clean and augmented data have the same labels, thus the CSA loss, which requires pairs of different labels, cannot be applied directly.
Therefore, we propose \textit{feature shuffling}, which makes different label pairs by shuffling the indices of augmented data features within a mini-batch.
Our method is simple yet effective and highly practical in terms of ease of use and maintenance.
Figure~\ref{fig:overview} shows an overview of the proposed method.

We evaluated the proposed method on the CIFAR-100~\cite{cifar} and CUB datasets~\cite{cub}.
Experiments on several model architectures and typical data augmentations show that the proposed method improves robustness and training efficiency despite only adding loss.

The main contributions of this paper are as follows.
 \begin{itemize}
    \item We treat clean data as source distribution data and augmented data as out-of-distribution data, thus data augmentation can benefit from SDG methods.
    \item We propose feature shuffling, which allows for the introduction of CSA loss into data augmentation.
    \item Experiments show that the proposed method further increases the robustness of commonly used data augmentations and improves training efficiency without losing the simplicity of data augmentation.
 \end{itemize}

\section{Related Work}
\label{sec:rw}

\subsection{Data Augmentation}
Data augmentation is one of the most common techniques used during deep learning training which increases the data amount and diversity by transforming the data.
In the past, simple transformations, such as flip and crop, were used~\cite{resnet}, then the augmentation policies~\cite{autoaugment, randaugment} and mixing of multiple data~\cite{mixup, cutmix, augmix, augmax} have been well studied.
In particular, MixUp~\cite{mixup} and CutMix~\cite{cutmix}, despite their simplicity, have been incorporated into the training of state-of-the-art architectures~\cite{bai2021transformers} because of their various advantages that improve accuracy, robustness, confidence calibration, and transferability.
In terms of improved robustness, AugMix~\cite{augmix}, which mixes multiple augmented data, and AugMax~\cite{augmax}, which incorporates adversarial training into AugMix, outperform other methods.
Despite these successes, data augmentation increases the complexity of training and often requires more epochs.
In this paper, we show that data augmentation can benefit from improved robustness and training efficiency from the SDG method.


\subsection{Feature Alignment}
Feature alignment improves accuracy by modifying the intermediate output of the deep neural network.
In the context of self-supervised learning, contrastive learning~\cite{simclr, simclrv2} has been studied, and supervised contrastive learning~\cite{supcon}, which extends it to supervised settings, has recently been proposed.
They are used during pre-training and are made to work effectively by adding two techniques: adding a multilayer perceptron and adding data augmentation such as crop.
Therefore, data augmentation cannot benefit from these techniques because they require additional processing, such as fine tuning and architectural modifications.
In the context of SDG, CSA loss~\cite{csa} is proposed, which maps features with the same label closer together and features with different labels farther apart.
In this paper, we propose a feature shuffling to apply CSA loss into data augmentation.
Some studies combine data augmentation and feature alignment~\cite{wang2021learning, wang2022toward}, but these require specially designed data augmentation and cannot be applied to existing data augmentation.


\section{Proposed Method}
\label{sec:pm}
\subsection{Applying SDG Methods to Data Augmentation}
Let $x^{\mathrm{aug}}$ be the augmented images from clean images $x$ in the training dataset $D=\{ (x_i, y_i) \}_{i=1}^{n}$.
We treat data augmentation as SDG by considering $x$ as source distribution images and $x^{\mathrm{aug}}$ as out-of-distribution images.

\subsection{Contrastive Semantic Alignment Loss}
In general, a model $f$ is composed of a feature extractor $g$ and a classifier $h$.
Here, $g \colon X \rightarrow Z$ is the embedding from the input space $X$ to a feature space $Z$, and $h \colon Z\rightarrow Y$is the function to predict class probabilities $Y$ from the feature space $Z$. 
With this notation, $f = g \circ h$.

CSA loss improves the robustness of classification models by training the feature extractor $g$.
CSA loss $L_{\mathrm{CSA}}$ consists of semantic alignment loss $L_{\mathrm{SA}}$ and separation loss $L_{\mathrm{S}}$.
Semantic alignment loss encourages data of different distributions but with the same label to map nearby in the embedding space.
In contrast, separation loss adds a penalty, thus data with different labels in different distributions are mapped farther apart in the embedding space.
Each loss is the following equation.

\begin{align}
\label{eq:csa_sa}
& L_{\mathrm{SA}}(g) = \Sigma_{i,j} \frac{1}{2} \|g(x_i) - g(x_j^{\mathrm{aug}})\|^2, \\
\label{eq:csa_s}
& L_{\mathrm{S}}(g) = \Sigma_{i,j} \frac{1}{2} \max (0, m-\|g(x_i) - g(x_j^{\mathrm{aug}})\|)^2, \\
\label{eq:csa}
& L_{\mathrm{CSA}}(g) = L_{\mathrm{SA}}(g) + L_{\mathrm{S}}(g).
\end{align}
Here, $\|\cdot\|$ is the Frobenius norm, $y$ is a ground-truth label, and $m$ is the margin that specifies separability in the feature space, set to 1 in this paper.
$L_{\mathrm{SA}}$ is used for training when the labels of $i, j$ are the same and $L_{\mathrm{S}}$ is used when they are different.

\subsection{Feature Shuffling}
Data augmentation produces augmented images from clean images, and the labels for these image pairs are identical, thus $i=j$ in Equations~\ref{eq:csa_sa} and~\ref{eq:csa_s}.
At this time, $L_\mathrm{S}$ is not used for training because the labels $y_i$ and $y_j$ are the same.
To prevent this, we randomly shuffle the feature index $j$ in the mini-batch, resulting in $i \neq j$ and $L_\mathrm{S}$ being trainable.


\subsection{Overall Loss for Data Augmentation}
MixUp and CutMix, which augment image by mixing different image, use the following loss to train the model.

\begin{align}
\label{eq:mixup_loss}
L_{\mathrm{aug}} = \lambda_m L(f(x^{\mathrm{aug}}), y^\mathrm{a}) + (1-\lambda_m) L(f(x^{\mathrm{aug}}), y^\mathrm{b}),
\end{align}
where $L$ is the cross-entropy loss, $y^\mathrm{a}$ and $y^\mathrm{b}$ are the labels of the image before mixing, and $\lambda_m$ is the mixing ratio of the image.
The total loss is the following equation.

\begin{align}
\label{eq:mixup_total_loss}
L_{\mathrm{total}} = (1-\gamma) L_{\mathrm{aug}} + \gamma L_{\mathrm{CSA}}.
\end{align}
$\gamma$ is a parameter to balance losses.
Although $x^{\mathrm{aug}}$ has two labels, for simplicity we train the CSA loss with only $y^\mathrm{a}$ as the label.

AugMix and AugMax, which augment the image by mixing the augmented image, use the following loss to train a model.

\begin{align}
\label{eq:augmix_loss}
L_{\mathrm{aug}} = L(f(x), y) + \lambda_l L_{\mathrm{JSD}}(f(x), f(x^{\mathrm{aug1}}), f(x^{\mathrm{aug2}})),
\end{align}
where $\lambda_l$ is the trade-off parameter, $L_{\mathrm{JSD}}$ is the Jensen-Shannon divergence consistency loss with two augmented image, $x^{\mathrm{aug1}}$ and $x^{\mathrm{aug2}}$.
To take advantage of these two augmented image, we used them in the CSA loss calculations.
The total loss is the following equation.

\begin{align}
\label{eq:augmix_total_loss}
L_{\mathrm{total}} = (1-\gamma) L_{\mathrm{aug}} +  \frac{1}{2} \gamma (L_{\mathrm{CSA}}^{\mathrm{aug1}} + L_{\mathrm{CSA}}^{\mathrm{aug2}}). 
\end{align}

\section{Experiment}
\label{sec:exp}
We experimented on two datasets, several classification models, and a typical data augmentation and found the proposed method to be effective.

\subsection{Setup}
\label{sec:setup}
For the experiments we used the CIFAR-100 dataset~\cite{cifar} and the CUB dataset~\cite{cub}.
In the CIFAR-100 experiments, we used SGD optimizer with momentum 0.9.
All models were trained with a batch size of 128 for 200 epochs with a weight decay of 0.0005. 
The learning rate started as 0.1 and decreased by the cosine annealing learning rate scheduler. 
In the CUB experiments, we used Adam~\cite{adam} optimizer with momentum 0.9.
A pretrained model on ImageNet~\cite{imagenet} was fine-tuned with a batch size of 64, 100 epochs, and a weight decay of 0.0005.
The learning rate started as 0.0001 and decreased by the factor of 0.5 at every 10 epochs. 
ResNet~\cite{resnet}, WRN~\cite{wrn} (WRN40-2 for CIFAR-100 and WRN50-2 for CUB), and ResNeXt~\cite{resnext} (ResNeXt29 for CIFAR-100 and ResNeXt50 for CUB) were used as classification models.
For data augmentation, we used AugMix, which is highly effective in improving robustness.
The parameter $\gamma$ for the balancing of losses was set at 0.25 for CIFAR-100 and 0.05 for CUB.
We evaluated our method on standard accuracy (SA), which is the accuracy of a normal test set, and robust accuracy (RA), which is the average accuracy of an artificially corrupted test set~\cite{in-c}.
All experiments were performed three times, and we report the average values.

\subsection{Experimental Results}

Table~\ref{tab:main_result} shows the experimental results evaluated on the two datasets, with CSA loss added to AugMix.
In most experiments, the proposed method improves SA, and in all experiments the proposed method improves RA.
In particular, ResNet18 experiment on CIFAR-100 shows results that greatly improve SA by 1.56 points and RA by 1.64 points.
Some results on the CUB dataset show that the proposed method slightly reduces SA but maintains a higher than normal.
Despite the simplicity of the proposed method, which only adds loss and requires no additional overhead, the gains are significant.

\begin{table}[t]
\centering
\caption{
Accuracy of models trained by AugMix with CSA loss on the two datasets. 
The numbers in parentheses indicate the differences from that without CSA loss.}
\label{tab:main_result}
\resizebox{\linewidth}{!}{
\begin{tabular}{cccccccc}
\hline
CIFAR-100 & Metric & Normal & AugMix & AugMix w/Ours \\ \hline
\multirow{2}{*}{ResNet18} & SA(\%) & 77.54 & 77.17 & \textbf{78.72}\color{green}(1.56) \\
 & RA(\%) & 48.67 & 64.72 & \textbf{66.36}\color{green}(1.64) \\ \hline
\multirow{2}{*}{ResNet50} & SA(\%) & 78.74 & 79.70 & \textbf{79.99}\color{green}(0.29) \\
 & RA(\%) & 50.61 & 67.23 & \textbf{67.96}\color{green}(0.73) \\ \hline
\multirow{2}{*}{WRN} & SA(\%) & 76.63 & 77.86 & \textbf{78.23}\color{green}(0.37) \\
 & RA(\%) & 47.88 & 64.99 & \textbf{65.26}\color{green}(0.26) \\ \hline
\multirow{2}{*}{ResNeXt} & SA(\%) & 80.01 & 79.91 & \textbf{80.27}\color{green}(0.36) \\
 & RA(\%) & 48.33 & 66.25 & \textbf{66.55}\color{green}(0.30) \\ \hline
\\
 \hline
CUB & Metric & Normal & AugMix & AugMix w/Ours \\ \hline
\multirow{2}{*}{ResNet18} & SA(\%) & 77.72 & \textbf{78.39} & 78.29\color{red}(-0.10) \\
 & RA(\%) & 44.48 & 55.89 & \textbf{56.33}\color{green}(0.44) \\ \hline
\multirow{2}{*}{ResNet50} & SA(\%) & 82.02 & 82.92 & \textbf{83.49}\color{green}(0.57) \\
 & RA(\%) & 48.29 & 61.99 & \textbf{63.47}\color{green}(1.48) \\ \hline
\multirow{2}{*}{WRN} & SA(\%) & 83.48 & 84.20 & \textbf{84.64}\color{green}(0.44) \\
 & RA(\%) & 51.33 & 64.86 & \textbf{66.11}\color{green}(1.24) \\ \hline
\multirow{2}{*}{ResNetXt} & SA(\%) & 83.75 & \textbf{84.33} & 84.32\color{red}(-0.01) \\
 & RA(\%) & 53.01 & 65.05 & \textbf{65.88}\color{green}(0.83) \\ \hline
 
\end{tabular}
}
\end{table}

\subsection{Experiments with Other Data Augmentations}
To evaluate the generality of the proposed method, we experimented with MixUp, CutMix, and AugMax.
In the AugMax experiment, we followed AugMax's experimental setup and changed the batch normalization layer of the classification models to a Dual Batch-and-Instance Normalization layer~\cite{augmax}.
Table~\ref{tab:other_da} shows the experimental results.
In almost all combinations, the proposed method improves SA and RA.
In particular, the gain to CutMix is significant, with the proposed method improving RA by 2.29 points on the CUB dataset.
Experiments with MixUp on the CUB dataset show that the proposed method decreases SA by 0.72 points, but as can be seen from Table~\ref{tab:main_result}, the SA of the proposed method is 82.22, which is higher than the normal SA of 81.74.

\begin{table}[tb]
\centering
\caption{Accuracy of models trained by each augmentation with CSA loss on the two datasets. The results indicate the mean accuracy of the four architectures, with the exception of AugMax, which shows the accuracy of ResNet18. The numbers in parentheses indicate the differences from that without CSA loss.}
\label{tab:other_da}
\resizebox{\linewidth}{!}{
\begin{tabular}{ccccc}
\hline
 Dataset & Metric & MixUp & CutMix & AugMax \\ \hline
 \multirow{2}{*}{CIFAR-100} & SA(\%) & 80.30\color{green}(0.25) & 80.74\color{green}(0.24) & 79.94\color{green}(0.31) \\
   & RA(\%) & 54.49\color{green}(0.10) & 49.68\color{green}(0.92) & 68.46\color{green}(0.15) \\ \hline
  \multirow{2}{*}{CUB} & SA(\%) & 82.22\color{red}(-0.72) & 82.12\color{green}(0.47) & 73.57\color{green}(0.79) \\
   & RA(\%) & 61.30\color{green}(1.04) & 50.88\color{green}(2.29) & 53.10\color{green}(1.51) \\ \hline
\end{tabular}
}
\end{table}

\subsection{Experiments on Few-epoch Training}
Data augmentation requires more training epochs than usual \cite{tied_aug}.
To verify how much of this problem can be mitigated by our method, we experimented with the number of epochs described in Sec.~\ref{sec:setup} as 25\%, 50\%, and 75\%.
Table~\ref{tab:epoch_acc} shows the results.
The proposed method improves the training efficiency of data augmentation and achieves high accuracy with a small number of epochs.

\begin{table}[t]
\centering
\caption{
Accuracy of ResNet50 trained by AugMix with CSA loss at different training epochs on the two datasets.
The numbers in parentheses indicate the differences from that without CSA loss.
}
\label{tab:epoch_acc}
\resizebox{\linewidth}{!}{
\begin{tabular}{ccccc}
\hline
Dataset & Metric & $25\%$ & $50\%$ & $75\%$ \\ \hline
\multirow{2}{*}{CIFAR-100} & SA(\%) & 77.01\color{green}(1.04) & \color{red}78.66(-0.11) & 79.95\color{green}(0.13) \\
 & RA(\%) & 64.41\color{green}(0.44) & 66.17\color{green}(0.31) & 68.33\color{green}(0.54) \\ \hline
\multirow{2}{*}{CUB} & SA(\%) & 82.50\color{green}(0.52) & 83.36\color{green}(0.22) & 83.72\color{green}(0.53) \\
 & RA(\%) & 61.99\color{green}(1.15) & 63.61\color{green}(1.99) & 63.24\color{green}(1.16) \\ \hline
\end{tabular}
}
\end{table}

\subsection{Ablation Study}

The proposed method optimizes $L_{\mathrm{SA}}$ and $L_{\mathrm{S}}$ by feature shuffling.
Table~\ref{tab:each_loss} shows the comparison results with the case where only $L_{\mathrm{SA}}$ is optimized without feature shuffling.
The results show that with feature shuffling outperforms those without it.
Optimizing $L_{\mathrm{SA}}$ only improves RA, but significantly decreases SA, indicating that $L_{\mathrm{S}}$, which increases the distance between features with different labels, is more beneficial for data augmentation.


\begin{table}[t]
\centering
\small
\caption{Accuracy of ResNet50 trained by AugMix with each loss on the CUB dataset. The numbers in parentheses indicate the differences from that without each loss.}
\label{tab:each_loss}
\begin{tabular}{cccc}
\hline
 Metric & CSA   & $L_{\mathrm{SA}}$ & SupCon  \\ \hline
 SA(\%) & 83.49\color{green}(0.57) & 80.90\color{red}(-2.02) & 83.17\color{green}(0.25) \\
 RA(\%) & 63.47\color{green}(1.48) & 62.72\color{green}(0.73) & 61.70\color{red}(-0.28) \\ \hline
\end{tabular}
\end{table}

\subsection{Comparison with Other Feature Alignment Loss}
Although Supervised Contrastive Learning (SupCon) is a promising feature alignment method, it is not suitable for data augmentation because SupCon is a pre-training method and requires two techniques to be effective (adding a multilayer perceptron layer and adding data augmentation such as crop).
Table~\ref{tab:each_loss} shows the accuracy when only the loss of SupCon is applied to data augmentation training.
The results show that the SupCon loss is inferior to the CSA loss.
The SupCon loss also results in lower RA than the standard AugMix.
We suspect that this result is due to the different usage conditions in which SupCon is most effective.

\subsection{Hyperparameter Sensitivity.}

\begin{figure}[t]
\centering
\includegraphics[width=1.0\linewidth]{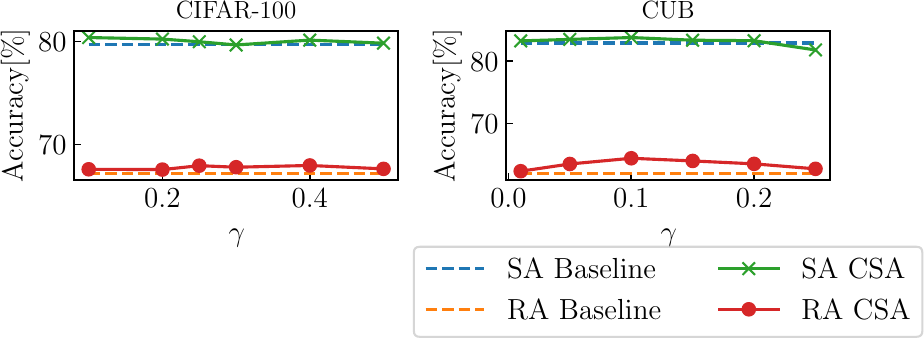}
\caption{Hyperparameter sensitivity of ResNet 50 with AugMix. We plot the accuracy as we vary the $\gamma$ from 0.1 to 0.5 for CIFAR-100 and from 0.01 to 0.25 for CUB.}
\label{fig:hypara_sensitivity}
\end{figure}

The proposed method has a hyperparameter $\gamma$ that balances the losses.
The relationship between hyperparameters and accuracy is shown in Figure~\ref{fig:hypara_sensitivity}.
The larger $\gamma$ tends to lower SA on the CUB dataset.
However, the proposed method consistently improves RA regardless of $\gamma$.

\section{Conclusion}
\label{sec:conc}

In this paper, we treat data augmentation as SDG.
To solve the training efficiency problem of data augmentation, we apply the SDG method, CSA loss, to data augmentation.
Training for CSA loss requires data pairs with different labels, and to accomplish this, we propose feature shuffling, which randomly shuffles feature indices.
The experimental results show that our method improves the robustness and training efficiency of data augmentation despite only adding loss.
An ablation study also shows the effectiveness of feature shuffling.
We hope that this study will be a baseline for incorporating SDG methods into data augmentation  and will lead to further development in this research area.

\vfill\pagebreak
\clearpage

\bibliographystyle{IEEEbib}
\bibliography{strings,refs}

\end{document}